# Cluster Workload Allocation: A Predictive Approach Leveraging Machine Learning Efficiency

**Leszek Sliwko**[1], **IEEE Professional**
[1]University of Westminster, London, W1B 2HW, UK

Corresponding author: Leszek Sliwko (e-mail: lsliwko@gmail.com, orcid: 0000-0002-1927-8710).

This work was supported in part by the AWS Activate Founders Program 2021 grant for meta-analyzer.com development.

**ABSTRACT** This research investigates how Machine Learning (ML) algorithms can assist in workload allocation strategies by detecting tasks with node affinity operators (referred to as constraint operators), which constrain their execution to a limited number of nodes. Using real-world Google Cluster Data (GCD) workload traces and the AGOCS framework, the study extracts node attributes and task constraints, then analyses them to identify suitable node-task pairings. It focuses on tasks that can be executed on either a single node or fewer than a thousand out of 12.5k nodes in the analysed GCD cluster. Task constraint operators are compacted, pre-processed with one-hot encoding, and used as features in a training dataset. Various ML classifiers, including Artificial Neural Networks, K-Nearest Neighbours, Decision Trees, Naive Bayes, Ridge Regression, Adaptive Boosting, and Bagging, are fine-tuned and assessed for accuracy and F1-scores. The final ensemble voting classifier model achieved 98% accuracy and a 1.5-1.8% misclassification rate for tasks with a single suitable node.

## I. INTRODUCTION

Cloud computing represents a firmly established technological advancement that enables users to collectively leverage resources on an unparalleled scale in the realm of information technology. Functioning through an interconnected network of individual servers, cloud systems efficiently handle interlinked tasks simultaneously across multiple environments. In comparison to individual computers with comparable computing capabilities, clouds typically present a more economical solution due to extensive resource sharing and simplicity of deploying pre-configured containers such as Docker images. An example of this is Amazon Web Services' pricing for their EC2 fleet, which can be as low as 2-3 cents per hour per box [1].

The considerable physical scale of these systems enables the involvement of thousands of machines, highlighting scalability and extensive reach of this ground-breaking technology. Despite this, the opportunities and substantial computational power provided by this technology pose their own set of challenges, in particular concerning the orchestration of computing cells and the effective balancing of incoming tasks and deployed services.

Cluster orchestration is the automated handling and improvement of cloud resources and services. It includes tasks like setting up, adjusting, and keeping track of different parts of cloud systems, such as virtual machines, storage, and networking. Its purpose is to make deploying and managing complex cloud setups both easier and more efficient, ensuring the better use of resources, scalability, and flexibility in cloud operations. At its core, cluster orchestration is an automated load balancing system which ensures the uninterrupted operation of services and tasks for users. While the majority of these systems are proprietary, with only limited details about their workings having been published, incidences of Open-Source systems do exist, for example Kubernetes [8], Apache Mesos [53], and Docker Swarm [54].

When committing a task or service to a cluster, users specify required thresholds for resources such as memory and the number of cores. Additionally, users can specify constraints for the task, limiting the nodes on which a given task can be executed. Depending on the implementation, such constraints can take the form of key-value pairs, matchers, or selectors (e.g., in Kubernetes). One of the most advanced constraints is the use of operators in Google's Borg cluster scheduler [2][3], where users can utilise simple functions such as greater than, less than, define inclusions and exclusions [35].

The focus of this research is to analyse tasks whose placement is limited by these kinds of constraints, and to predict the required effort in allocating a given task to a computing cell. These predictions help the scheduling logic to allocate more computing time or even to use a different category of algorithms once such a task is detected. Due to the substantial amount of data to examine and inherent complexity of such structures – specifically, more than 900,000 tasks with tens of different constraints identified – Machine Learning (ML) models were deployed to learn and analyse the data.

The main contributions of this work are summarised as follows:

1. This paper is a novel attempt to improve task allocations by analysing and modelling execution constraints (also referred to as 'node-affinity' in Kubernetes clusters [8]). In previous research, task execution constraints were often overlooked, with the focus primarily on optimising resource utilisation or service quality attributes such as throughput and latency. At best, task-node affinity and constraint logic were simplified to checking whether a given node could support a particular task (an approach to which the author of this paper also admits [36][39]);
2. Analysing real-world workload traces from Google Cluster Data (GCD) [4] and identifying approximately 2% of the total tasks with constraint operators as hard to allocate among cluster workload traces. Further splitting them into twenty-six groups A-Z, which describe how many nodes a given task is able to be allocated to, and focusing on identifying two groups: Group A (containing tasks that can be scheduled only on a single node) and Group C (containing tasks that can be scheduled on fewer than a thousand nodes);
3. Successfully selecting a number of ML classifier models capable of being trained on datasets with features spanning into thousands. Efficiently tuning them to achieve the best results, and then comparing the outcomes and suitability for practical deployment in a cluster scheduler. In the last step, the three best were selected based on their results, accuracy, and the ratio of misclassifications for Groups A and C, and combined into the final ensemble voting classifier.

The paper is organised as follows. Section 2 details related work with a focus on load balancing and task scheduling utilising ML algorithms. Section 3 provides justification for the research, describes sources of workload traces data, and gives examples of constraint operators. Section 4 presents the challenges of preparing the dataset for training ML models, and also discusses the ways those challenges were overcome. Section 5 presents selected ML algorithms, details the process of how they were tuned, and also presents the outcomes of predictions. Section 6 contains a brief summary, details limitations of the research approach, and sets the scope of future work.

## II. RELATED WORK

The focus of this research is task scheduling in a cluster environment. The methods of task scheduling have been a subject of extensive research for a considerable period, with numerous effective solutions designed, implemented, and deployed globally in real-world scenarios. Some noteworthy mentions include:

- Microsoft's Apollo design is finely tuned for high task churn, handling over 100,000 scheduling requests per second on a cluster of around 20,000 nodes during peak times. Using a set of per-job schedulers called Job Managers, Apollo efficiently manages tasks within job entities, typically comprising short-lived batch jobs. Each node operates with its own Process Node managing local scheduling decisions and prioritising smaller tasks for immediate execution while larger tasks wait for available resources [5].
- The Fuxi (named after the Chinese deity '伏羲') employs a distinctive approach by matching newly-available resources with the backlog of tasks, rather than matching tasks to available resources on nodes. This technique has enabled Fuxi to achieve exceptionally high resource utilisation rates, reaching 95% for memory and 91% for CPU. Supporting Alibaba's workload since 2009, Fuxi scales to approximately 5,000 nodes [6].
- Google's Borg employs multiple schedulers operating in parallel, sharing the state of available resources with optimistic concurrency control. Conflicts arise when multiple schedulers allocate jobs to the same resources, resulting in all involved jobs being returned to the queue [30]. When allocating tasks, Borg's scheduler evaluates available nodes and selects the most suitable one. Initially, Borg utilised a variation of the Enhanced Parallel Virtual Machine algorithm [58], ensuring fair task distribution but leading to increased fragmentation and difficulties accommodating large jobs that required significant node resources. In contrast, the best-fit strategy tightly packs tasks, potentially causing excessive pre-emption of other tasks on the same node, particularly during resource miscalculations or application load spikes. Borg's current scheduling model is a hybrid approach aiming to minimise resource usage gaps [2].
- Historically, Omega originated as a spinoff from Borg scheduler. Despite Borg's ongoing enhancements, Google chose to develop Omega anew to tackle head-of-line blocking and scalability issues. Omega introduced innovations like centralised cluster state storage using a Paxos-based system, resolving conflicts with optimistic locking concurrency control. This allowed Omega to run multiple schedulers simultaneously, boosting scheduling throughput. Many of Omega's innovations have since been integrated into Borg [57].

The details of the algorithms employed in these cases remain proprietary solutions of the respective companies, and as such their workings are not disclosed. Nevertheless, widely adopted open-source alternatives such as Docker Swarm [7] and Kubernetes [8] provide accessible alternatives in the realm of task scheduling.

In addition, in recent years, the integration of ML algorithms into cluster scheduling and load balancing has garnered significant attention due to its potential to optimise resource allocation and enhance system performance. In this section, a concise overview of notable research contributions in the realm of load balancing employing ML is presented. This research is primarily, although not exclusively, centred around cloud computing. It should be noted that the list below emphasises more recent contributions due to the extensive number of relevant publications.

In [9], researchers are exploring the potential of integrating ML-based approaches into software-defined networking architectures to enhance network resource utilisation and overall performance. Alongside other algorithms, the authors present a classification of AI-based load balancing methods such as Bayesian Networks, K-means, neural networks and multiple linear regression. The paper itself serves as a thorough introduction to load balancing strategies in general, covering the period 2017-2023 – a period which coincides with heightened interest in ML, as indicated by Google Trends. Hence, the content is fairly recent, encompassing the timeframe during which much of the research on ML-based workload scheduling occurred. Dominant trends observed included the transition of research from conventional machine learning methods such as regression, to more advanced deep learning architectures like Neural Networks [10][11][12] and Long Short-Term Memory models [13].

[14] categorises workload traces from various applications and network infrastructure usage. This work discusses workload features, models and techniques for their evaluation which are applied in resource planning. The paper offers a good selection of publications from 1996 onwards, covering web traffic, mobile app traffic, video service traffic and – most interestingly for this research – cloud workloads, including various virtualisation services like virtual desktops.

[15] is another noteworthy survey of the methods of resource scheduling focusing in particular on deep learning and reinforcement learning. The paper reviews a large number of publications spanning from the early 2000s to 2021. It divides the listed approaches into reinforcement learning, deep learning, and a combination of these two. In addition, the authors provide an insightful section discussing the challenges and future prospects in cloud computing scheduling.

[16] investigates the challenge of ensuring reliable resource provisioning in joint edge-cloud environments, analysing a range of technologies, mechanisms, and methods aimed at enhancing the reliability of distributed applications across diverse and heterogeneous network environments. The survey is structured around three load balancing problems: workload characterisation and prediction, placement of components, and managing a running application. Special attention is given to the utilisation of machine learning frameworks.

[18] examines a number of papers utilising ML algorithms such as linear regression, random forest classifiers, neural networks, convolutional neural networks, and long-short term memory-recurrent neural networks.

[12] proposes a scheduling scheme based on Neural Networks with elements of Reinforcement Learning. The authors focused on developing an encoder to analyse and provide a vectorised form of the workflow scheduling problems. Parameters representing the workflow's structure, task characteristics, nodes, and performance models, amongst other factors, are used to estimate task execution time.

[17] examines statistical and machine learning methods for forecasting workload using past values. Among proposed models, it uses machine learning methods such as Bayesian networks, neural networks, SVMs, and decision trees.

[19] proposes a load balancing framework that utilises deep learning-based techniques to generate optimal schedules, before comparing the results obtained with other load balancers developed by the same authors.

[20] developed a model wherein hash functions in load balancing mechanisms are substituted with deep learning models. These models are trained to effectively map various distributions of workloads to servers in a uniform manner. All testing was done with the help of CloudSim.

In [21], the authors proposed a load balancing strategy primarily based on SVM and K-means clustering techniques. The SVM is used to categorise entity activity based on their size, with K-means clustering then employed to form groups of Virtual Machines based on their CPU and approximated RAM utilisation.

In [22], the authors acknowledge the challenges of applying and deploying machine learning technology in software-defined networking architectures. They propose using a Bayesian network to predict the degree of load congestion and employ a reinforcement learning algorithm to make optimal action decisions.

In [23], the authors propose heuristic and load balancing models based on algorithms such as multiple linear regression, random forest walk, and AdaBoost techniques to dynamically determine the processing unit for each incoming query.

[10] introduces a workload prediction model that employs neural networks and a differential evolution algorithm capable of learning and automatically selecting the most suitable mutation strategy and crossover rate.

[24] presents an interesting approach to load balancing, wherein a machine learning algorithm dynamically selects the most suitable load balancing strategy based on previously collected information about the application. The paper evaluates a variety of ML algorithms and compares their suitability for this task.

[46] explores the potential and limitations of real-time learning of job runtime properties by sampling and scheduling a small fraction of the tasks of each job, and then estimating the whole job runtime properties based on the results of those initial tasks. While this approach does not utilise ML models, the authors evaluate their research using three distinctive cluster workload traces and achieve substantial reductions in the average job completion time.

[52] deploys a hybrid model utilising a forecasting algorithm based on a recurrent LSTM neural network model with Particle Swarm Intelligence and Genetic Algorithm for dynamic workload provisioning in cloud computing. The algorithm optimises metrics such as throughput, resource utilisation, latency, response times, and costs in cloud environments. Noteworthy - the authors utilise GCD workload traces to validate the effectiveness of their approach.

[56] suggests using a deep learning model to monitor and predict the parameters in the workload of a clustered system orchestrated by Kubernetes. The framework then uses those parameters to fine-tune scheduling algorithms. The paper explains the technical details of extending Kubernetes modules for AI-assisted modules; however, the authors do not disclose which ML methods were deployed.

[59] enhances the native Kubernetes scheduler with a reinforcement learning module based on a policy optimisation with the least response time algorithm. The solution dynamically selects the most suitable worker nodes with the shortest response time according to the current cluster load and pod state. This research is noteworthy as the authors tested their solution by scheduling five-hundred random tasks on a live Kubernetes cluster and demonstrated the effectiveness of their solution with even allocation of the workload across the cluster.

The above publications address the challenges of scheduling a workload on clustered systems with the help of ML methods. However, they rarely utilise accurate and detailed simulations of the distributed environment, as the effort required to implement and set up a sophisticated cloud simulator is not trivial. Additionally, the research mainly focuses on task allocation based on available resources or service quality attributes. The task-node affinity and constraints logic are usually simplified or outright ignored. This paper is the first attempt to improve task allocations based on analysing and modelling execution constraints.

## III. RESEARCH MOTIVATION

Although the field of task scheduling has been explored extensively, there remain significant niches that hold potential for improvement. This aim of this research is to enhance a specific aspect of task scheduling, namely predicting the required effort in allocating a given task to a computing cell, which in turn will help the scheduling logic to allocate more computing time or even use different category of algorithms once such a task is detected.

### A. WORKLOAD TRACES

Theoretical investigations into the Cloud's resources model [5][6][10][30][57][7][46] have highlighted the immense complexity of the workings of the Cloud environment. It became obvious that statistical data and dry analysis alone were insufficient for the research to progress, and that a more detailed and concrete approach was required, such as an analysis of the actually recorded Cloud workload.

For the purpose of this research, the following publicly available workload trace archives were considered:

- GCD project, which provides detailed traces spanning one month (May 2011) from a 12.5K-node network (clusterdata-2011) and eight different Borg cells for one month (May 2019) (clusterdata-2019). Available on GitHub (github.com/google/cluster-data), these traces encompass statistics such as CPU usage, memory consumption, disk I/O operations (limited to the first two weeks due to configuration changes), and network speed. In April 2020, the GCD archive was updated with newer traces from eight Borg cells taken from May 2019 [25];
- The Grid Workload Archive [26], located at Delft University of Technology in the Netherlands, hosts workload traces from nearly a dozen grid systems. The majority of these traces include data on CPU usage, memory usage, and disk I/O operations;
- The Parallel Workloads Archive [27] features over thirty workload logs gathered worldwide, spanning from 1993 to 2012. These traces have been meticulously cleaned to remove anomalies and data errors;
- The MetaCentrum Workload Log and CERIT-SC Grid Workload [28] archives contain datasets derived from TORQUE workload traces utilised in the Czech National Grid Infrastructure, comprising 22 clusters with 219 nodes housing 1494 CPUs;
- The Yahoo! M45 Supercomputing Project [29] has made available the workload traces of its 4k-node Hadoop cluster to select universities for academic research purposes.

All the repositories listed, excluding Yahoo!'s workload logs, are open for research without legal constraints. In this study, real-world workload traces from the GCD project are utilised due to their exceptional quality. These traces are comprehensive, containing few anomalies, and are also well-documented, making them ideal for analysis. Google's diverse range of services results in a varied spectrum of computational demands. Search engines, in particular, require substantial computation, with each query processing large volumes of data, guided by algorithms like PageRank. The search process also involves services like Spell Checker and Ads, which are crucial for Google's revenue.

Google engineers have prioritised a throughput-oriented framework. Analyses of GCD workloads reveal that the 80th percentile of batch jobs are completed within 12 to 20 minutes, while tasks finish within thirty days. Batch jobs constitute the majority (approximately 80%) of all tasks, and typically exhibit fast turnaround times with short execution periods. In Google Cluster, their 80th percentile inter-arrival time ranges from 4 to 7 seconds.

Given this, a low-overhead, low-quality allocation algorithm is suitable for batch jobs. Services, conversely, operate for longer durations and often involve interactive elements, necessitating high-quality allocation to ensure optimal performance. In Google Cluster, less than 20% of all jobs have an 80th percentile inter-arrival time of between 2 to 15 minutes. However, services consume a significant portion of system resources, accounting for approximately 55-80% of all resources in Google Cluster [30]. These figures align with findings from similar analyses of cluster workload traces from other companies such as Yahoo! [29], Facebook [31], and Google [32][33][34][35].

This mixed workload pattern provides an excellent testbed for designing a flexible and scalable scheduler capable of managing high workloads across any number of nodes. While other designs may focus on handling short-lived batch tasks or long-term services, Google's global operations ensure non-cyclical workload patterns, unlike localised data centres, thereby facilitating continuous operation.

The most notable update between May 2011 and April 2020 was the inclusion of power utilisation traces (powerdata-2019) for 57 power domains in Google data centres, encompassing the eight cells highlighted in clusterdata-2019 [25][84]. As part of this research, the AGOCS tool was adapted to the clusterdata-2019 data format; however, during simulation trials, it became evident that these traces contain numerous anomalies, making them challenging to use as a replay source. In particular: (i) the reported timings of work trace events are not accurate, and task submission or update events are frequently timed before task termination (i.e., eviction, failure, finishing, loss, termination, etc.), and (ii) certain tasks lack eviction or failure events, rendering AGOCS unable to remove them from the pool of running tasks. Consequently, this research opted for workload traces from 2011 (clusterdata-2011), which have been successfully simulated in prior studies [36][39].

### B. CLOUD SIMULATION

Assessing the performance of distributed applications and services becomes challenging when the access to existing cloud environments is limited. Simulation offers a viable approach to tackle this challenge. Whilst the available cloud simulators can effectively model high-level infrastructure parameters such as nodes, tasks, and basic dependencies, they are unable to provide realistic and detailed system traces, including application and canonical memory, local and remote disk space, disk I/O, cycles per instruction, etc. as used in real-world cluster environments [35]. However, accurately characterising those dependencies is of the utmost importance when modelling cloud workloads, especially when examining the lifecycles of tasks.

Previous work on simulating cluster workload resulted in the creation of the Accurate Google Cloud Simulator (AGOCS) [36] – a high-fidelity cloud workload simulator based on parsing real-world workload traces. At its core, AGOCS is a replay tool which can utilise a number of different workload formats, parse data, and analyse the workload dependencies so as to gather a range of statistical metrics. The simulation tool can work in two modes:

- Replay Mode, where it is actively timing the loaded workload traces according to stored instructions from cluster scheduler or schedulers. For example, GDC clusterdata-2019 workload traces [84] contain instructions from the primary (default) scheduler and secondary (batch) scheduler optimised for running jobs with low internal charges and no associated Service Level Objectives [37]. This setup is designed, primarily, to gather data-level statistics,
- Simulation Mode, in which the system ignores the input of scheduling instructions and reads the input from the provided function (or functions). This mode was mainly utilised to run extensive research over meta-heuristic based scheduling [38] and a distributed negotiation-based scheduling framework [39].

Initially, the design of AGOCS focused on providing a testbed simulator for a range of scheduling strategies, primarily aimed at optimising resource utilisation. However, for the purposes of this research, the AGOCS simulator has been additionally modified to generate outputs for all new tasks along with their corresponding constraint operators. It has also been adjusted to output the list of nodes, including their attributes, at five-minute intervals of simulation time.

The AGOCS is implemented using Scala programming language, and utilises a variety of additional packages including Scala's Parallel Collections, Apache's Lang-Commons, Math, I/O, and CSV libraries, Google's Guava, Kryo Serialization, and Akka Actors frameworks. The tool has been profiled and optimised to function with Scala 2.13. Detailed design and implementation for AGOCS is provided in [36].

### C. CONSTRAINT OPERATORS

This investigation focuses on tasks with defined constraint operators. The GCD workload traces contain extensive information about scheduled tasks, including their state in the lifecycle graph, required and utilised memory and CPU resources, disk I/O operations (limited to the initial two weeks due to a change in log configuration thereafter), network speed, and the constraint operators associated with task scheduling.

The actual categories of attributes were obfuscated in the GDC workload traces. However, schema definition provided within the GDC discloses that the task constraints not only check for the existence of (or exclude) a given attribute, but can also specify minimum or maximum values (or both) for numeric values. For example, a given task (or batch of tasks) could be run only on nodes which have an external IP address, or which have a particular version of system kernel. Nodes in the GCD workload traces define 67 unique attributes.

At any given time, there can be between 14% and 38% of tasks on cluster with constraint operators. During simulation, i.e. thirty days of GCD workload traces, 23.29% – on average – of scheduled tasks had at least one constraint operator. It should be noted that, almost exclusively, all tasks with constraint operators were long-running services, i.e., tasks running for more than twenty minutes. Figure 1 presents the split between tasks with and without constraint operators.

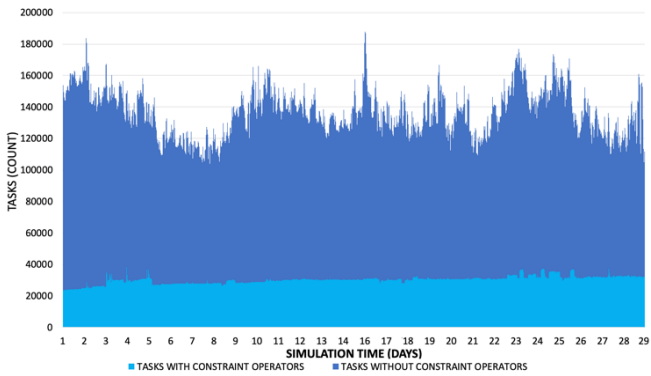

**FIGURE 1.** Tasks in five-minutes samples (GCD simulation time)

The GDC schema specifies four logical constraint operators originally coded as numeric values [35], which are translated into:
- Equal operator applies to both numeric and text values. The attribute must be present on the node and possess a value identical to the specified constraint value. If no value is specified, the attribute can be empty;
- Not-Equal is applicable to both numeric and text values, this operator requires that the attribute is either absent from the node's attribute list or has a value different from the specified constraint value;
- Less-Than, an operator which is relevant only to numeric values and which demands that the node's attribute value be less than the specified constraint value;
- Greater-Than, an operator which is specifically designed for numeric values, and which mandates that the node's attribute value must be greater than the specified constraint value.

LISTING 1
SAMPLE COMPACTION SCALA FUNCTION

```
def compactGreaterConstraints(taskConstraints: TaskConstraints) = {
  val greaterEqualThanAttributeConstraintGroups =
    taskConstraints.attributeConstraints
      .collect { case c: GreaterEqualThanAttributeConstraint => c }
      .groupBy(_.attributeName)

  val multipleGreaterEqualThanAttributeConstraintGroups =
    greaterEqualThanAttributeConstraintGroups
      .filter(_._2.size > 1).map(_._2)

  val flattenGreaterEqualThanAttributeConstraintGroups =
    multipleGreaterEqualThanAttributeConstraintGroups
      .map { group => group.maxBy(_.constraintValue) }

  TaskConstraints(
    taskConstraints.attributeConstraints
    -- multipleGreaterEqualThanAttributeConstraintGroups.flatten
    ++ flattenGreaterEqualThanAttributeConstraintGroups
  )
}
```

A given task might have many constraint operators on different attribute values. Furthermore, several operators on the same attribute value, for example a task, might have operators \${A} <> 0, \${A} == 4 and \${A} >= 0. Table 1 presents samples of tasks' constraints.

TABLE I
SAMPLE CONSTRAINT OPERATORS

| Task id (with job index) | Constraint operators |
|---|---|
| 5832654850-0 | \${E} >= 0, \${R} >= 0, \${D} == |
| 515042969-1 | \${N} <> 'ho', \${N} <> 'hp', \${D} == |
| 5889934721-6 | \${E} >= 0, \${AM} >= 2 |
| 4028922835-42 | \${W} >= 4, \${W} < 14 |
| 515042969-9 | \${N} <> 'ho', \${N} <> 'hp', \${N} <> 'hq' |
| 17109330-135 | \${A} <> 1, \${A} <> 0 |
| 1342043214-9 | \${AK} <> 'qe', \${AK} <> 'qg', \${AK} <> 'qh' |
| 2902878525-107 | \${W} >= 0, \${W} < 3 |

As part of the process of reading and parsing GDC files, the AGOCS framework collapses operators and introduces new ones:
- The Not-Equal-Array compacting operator is formed by consolidating a set of Non-Equal constraints into an array. It is versatile and applicable to all types of values, including both numeric and text;
- The Between operator is generated by pairing the Less-Than, Greater-Than, and Not-Equal (only for numeric values) operators, extracting the minimum and maximum values, and then creating a new constraint operator.

The sample compaction Scala function is presented in Listing 1, while Table 2 demonstrates examples of how the constraint operator collapse works. The constraint operators are grouped by the constraint variable name and then compacted into a single constraint operator. Where this isn't possible, an error is reported and such tasks are ignored. However, during simulation, only a tiny fraction of constraint operators were found to be invalid, and upon detailed examination, all of these were from failed executions of batch jobs. Although a task might have many constraint operators, collapsing several constraint operators into one ensures the task should have only one constraint operator per attribute.

The significance of this functionality is demonstrated later in the paper where encoding categories process, i.e. converting the constraint operators into zero-one vectors, takes place. As a side note, the constraint operators for the tasks might change during execution. However, for the purpose of the research (and training ML), the last recorded task constraints were utilised.

TABLE 2
SAMPLE CONSTRAINT OPERATORS COMPACTIONS

| Input constraint operators | Collapsed constraint operator |
|---|---|
| ${A} <> 'x'<br>${A} <> 'y'<br>${A} <> 'z' | ${A} <> 'x','y','z'<br>(operators are compacted into a new Non-Equal-Array operator) |
| ${B} >= 0<br>${B} <> 0 | ${B} >= 1<br>(operators are compacted) |
| ${C} >= 3<br>${C} < 5 | 5 > ${C} >= 3<br>(operators are compacted into a new Between operator) |
| ${D} = 'x'<br>${D} <> 'y'<br>${D} <> 'z' | ${D} = 'x'<br>(Not-Equal operators are removed as Equals operator is restrictive) |

### D. GROUPPINGS OF TASK ALLOCATIONS

The cluster orchestration software is responsible for not only assigning tasks to nodes but also for monitoring task health, sustaining the requested number of replicas, facilitating load balancing for network traffic, securing storage for secrets, and ensuring the coordinated management of related tasks as a group (referred to as gang-scheduling), amongst other responsibilities. In essence, a bustling cluster environment demands substantial computing power to uphold the continuous operation of all its components.

In prior research [39], the publicly available GCD was utilised to simulate the traffic of a real-life cluster scheduler. Further research saw the development of a distributed scheduling system capable of managing very large computing cells, comprising over 100,000 machines. One of the challenges in developing such a system were the tasks with logical constraint operators, and how they could be permitted to run only on a very limited number of suitable nodes, i.e. nodes whose attributes are matched by the operators mentioned.

As detailed in [39], the research initially aimed to enhance Google's Borg allocations. However, it became apparent that Google's engineers had already implemented a stellar design, and significant improvement in throughput was unattainable through this research. Consequently, the focus shifted to designing a solution capable of accommodating multiple sizes of Borg computing cells, i.e. as twelve thousand nodes, as recorded in the GCD workload traces. The paper introduced a decentralised agent-based load balancer along with a novel negotiation of tasks' re-allocations protocol. Using this approach, task allocation and scheduling logic were cooperatively executed on all nodes in the cluster rather than on a single machine (or a few interconnected machines). This decentralisation enabled the cell size to scale up to eight times the size of a Borg computing cell, and in the final tests 100k nodes were simulated during a month-long simulation.

One of the unsolved challenges in this research were specialised tasks with restrictive constraint operands, i.e., approximately 10-15 out of every 10k tasks presented constraints which restricted the execution of a task to a very limited number of nodes. In extreme cases, only a single node was suitable for processing these tasks. The negotiation protocol presented in [39] included a 'forced-migration' flag, which mandated the target node to reallocate currently running tasks. While this solution was functional, it was not ideal, as it caused unnecessary workload spikes and prematurely offloaded other tasks from the target node. This is particularly significant when only a limited number of nodes in the cluster are capable of executing such tasks.

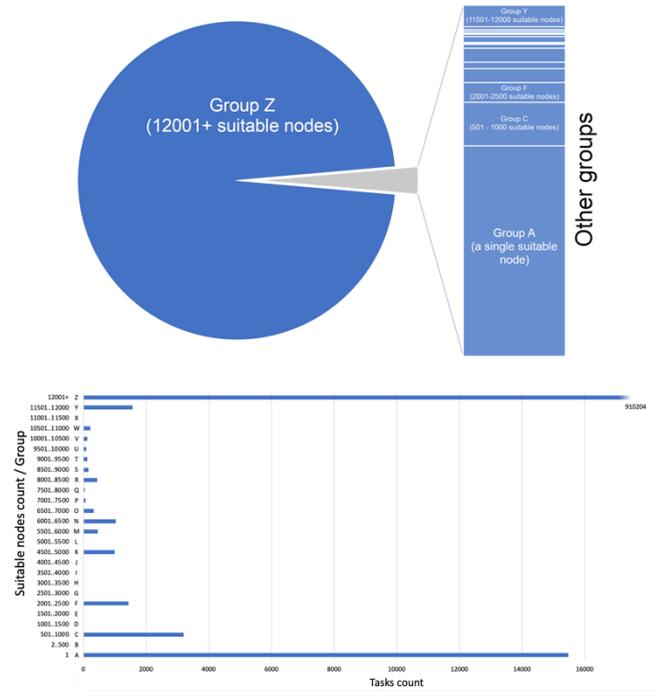

**FIGURE 2. The distribution of tasks based on the number of nodes they are suitable for (Groups A-Z). Group Z contains 97% of all tasks.**

Figure 2 illustrates the distribution of tasks based on the number of nodes they are suitable for, together with their designated Group A-Z. In the researched sample of 935.9k tasks with constraint operators (out of 48.4 million total scheduled tasks over thirty days contained in the GCD workload traces), the majority of tasks, i.e. 910.2k, could be allocated to virtually any cluster node. However, 15.4k tasks could be allocated to precisely one node – this set is referred to as Group A; the next sections of the paper will explain the significance of this group. Additionally, 3.2k tasks could be allocated to fewer than 1,000 nodes – these tasks are referred to as Group C. There were no tasks in Group B, i.e. tasks which could be allocated to between two and five hundred nodes.

This research aims to address this issue using the following strategy: when the rare node reaches its task capacity, the only option is to forcibly terminate (or otherwise remove) tasks running on it in order to make space for a new task, which entails limited suitable nodes. This process incurs costs both in terms of resources and the time needed to reinitialise offloaded tasks. In the simulation [39], this occurred in 6-8% of all task allocation attempts, with the majority resulting from task reallocation collisions where two or more tasks are being migrated to the same node.

The count of suitable nodes is a good predictor as to how much effort is required a given task is to allocate. This is especially applicable to Groups A and C, which collectively account for approximately 2% of the total tasks involving constraint operators. These groups consist of tasks that can only run on a very limited number of nodes. Therefore, these two groups are the primary focus of this research.

By harnessing recent advancements in ML frameworks, it becomes possible to predict which tasks might require such a node, facilitating the strategic allocation of tasks to these rare nodes only when absolutely necessary. The research question emerged out of this situation, namely: **Can the number of suitable nodes for a given task be predicted based on its constraint operators?**

The AGOCS was therefore modified to compute a list of all the nodes which could accommodate a given task based on specified constraint operators and attributes of the nodes. During the replaying of GCD workload traces, the nodes' attributes are occasionally updated, nodes are removed and new nodes are added. As such, the simulation re-computes the list upon every node update. However, the node changes are relatively rare event (out of 12.5k nodes, only 17 nodes were removed over the period of thirty days.

## IV. TRAINING DATASET

To evaluate the research question thoroughly, a tailored program was developed, harnessing the capabilities of the SciKit-learn [40] (SciKit-learn package version 1.4.1.post1 for Python 3.12.2), Pandas [41] (pandas package version 2.2.2), and NumPy [42] (NumPy package version 1.26.4) frameworks. Integrating these frameworks ensured a synergy between efficient data manipulation (Pandas and NumPy) and the application of a diverse range of artificial intelligence algorithms (SciKit-learn). Pandas and NumPy are auxiliary libraries in the realm of data science and programming which play an important role in streamlining and enhancing the manipulation of data and arrays. These libraries offer a rich set of functions and methods that significantly simplify tasks such as data cleaning, transformation, and analysis, thereby providing a robust foundation for efficient and effective data handling. The SciKit-learn framework is a gold standard for AI-related research and offers a comprehensive suite of both supervised and unsupervised ML algorithms straight out of the box.

Using the first approach, several attempts were made to model constraint operators such as greater than, between and not-equal with all possible integer and text values of attributes extracted from the workload traces of the nodes. However, the resulting datasets had lengthy feature lists (over 2,000 combined numeric and category-like features per data point). Furthermore, dimensionality reduction methods such as Principal Component Analysis and Latent Dirichlet Allocation (unsupervised) and Truncated Singular Value Decomposition (as implemented in SciKit-learn framework) were not yielding good results. Additionally, the data was sparse, thereby reducing the usefulness of many ML algorithms. To make the dataset more usable, several customisations and pre-processing steps were applied, which are described in the following paragraphs.

### A. CONSTRAINT OPERATORS' ENCODING

After conducting several experiments using different approaches on a partial dataset, a viable solution was identified. In this solution, presented in the Listing 2, the text representation of the constraint operator was treated as a category and encoded using a One-Hot Encoder pre-processor [55].

LISTING 2
ONE-HOT ENCODING OF CONSTRAINT OPERATORS

```
import pandas as pd

from sklearn.compose import ColumnTransformer
from sklearn.preprocessing import OneHotEncoder

# load data points
dataset = pd.read_csv('datapoint-tasks-raw.csv', low_memory=False)

X = dataset.iloc[:, 2:].values # columns 3-70 contain features
y = dataset.iloc[:, 0].values  # first column contains suitable nodes count

features_count = len(X[0]) # count features

# categorise features into zero-one vector
columns_list = [i for i in range(2, features_count)]
transformer = ColumnTransformer(
    transformers=[
      ('encoder', OneHotEncoder(drop='first'), columns_list)
    ],
    remainder='passthrough',
    sparse_threshold=0,
    n_jobs=-1  # use all cores
)

X_cat_encoded = transformer.fit_transform(X)
```

Unfortunately, when an attempt was made to encode the full dataset of 935.9k data points, the OneHotEncoder provided with the SciKit-learn package proved incapable of handling the huge volume of data, and ran out of memory on a testbed 2023 MacBook Pro Apple M2 Pro Chip with 12-Core CPU and 16GB RAM. Consequently, a custom encoder was developed in the Scala programming language using the Apache POI and Google Guava libraries, along with Java input streams. Several further low-level optimisations were implemented, such as dynamic unloading of Apache POI objects (specifically XSSFCell and XSSFRow instances) and caching variables with the help of Google Guava library's HashMultimap. To shorten computation time and to take advantage of multiple cores, the program was parallelised with the use of the Scala Parallel Collections library – since Scala 2.13, the above framework must be explicitly imported.

Aside from programming tweaks, the research replicated the SciKit-learn OneHotEncoder's feature of conditionally removing the first category in each feature dictionary (or eliminating the feature entirely if only one category was found). With the above optimisations, the length of the features array was set to 4404, including the requested memory and CPU time (normalised) values, resulting in an 8.28GB CSV file.

### B. TASKS' DATA POINTS COMPRESSION

Despite all the optimisations described in the previous section, the resulting dataset file with 935.9k data points still caused many ML algorithms to run out of memory, rendering them impractical. With the feature array substantially reduced, the next attempt aimed to reduce the size of the data points while preserving the information required for ML training.

However, because the constraint operators' values were converted into one-zero vectors, the research was able to take advantage of NumPy. NumPy is a library for the Python programming language, which extends support for large, multi-dimensional arrays and matrices, and is accompanied by an extensive collection of high-level mathematical functions designed to operate on these arrays. Of particular importance is that NumPy is written almost entirely in C and wrapped in Python for ease of use, thus excelling in computation speed and also exhibiting low memory usage, especially for numeric matrices such as the dataset under research.

On a related note, the SciKit-learn framework extensively utilises NumPy, as well as the BLAS and LAPACK libraries [43], which were initially implemented in Fortran programming language in 1979 and 1992 respectively. Both libraries have multiple implementations targeting various hardware platforms, including cuBLAS (NVIDIA GPU and GPGPU), rocBLAS (AMD GPU), OpenBLAS, and a recent implementation in the Apple Accelerate framework [44]. For instance, the Principal Component Analysis decomposition SciKit-learn implementation internally defaults to NumPy functions like 'numpy.searchsorted', and then performs decomposition by calling LAPACK low-level functions by name, such as 'sgesdd_lwork' and 'dgesdd_lwork' (the initial six-character segment of the function's name has its origins in Fortran 77's limitation). Please refer to the lapack.py source file for a full list of mapped functions.

In order to reduce the large number of data points, the design presented leveraged another feature of the examined workload traces, namely gang scheduling [45]. The task identifier in the GCD traces contains two values – the task id and index [35]. This is because Google's schedulers implement the gang-scheduling mechanism, ensuring that scheduling-related processes occur simultaneously across parallel system nodes and ensuring their simultaneous execution. In cluster-level gang-scheduling, as implemented at Google, all processes are typically either scheduled at once or not scheduled until conditions are met.

Examination of GCD workload trace events revealed that tasks associated with a given parent task id mostly shared constraint operators, as well as memory and CPU requirements. Occasionally, there are two or three distinct configuration groups for the same parent task id, understood as unique sets of memory and CPU requirements with the same constraint operators. Therefore, the 935.9k data points could be grouped into unique configurations per parent task id and one or more distinct configuration groups, with duplicate configurations removed, resulting in a reduction to 27.7k data points. With the data compression process completed as described above, the dataset CSV file was reduced to 245MB and was prepared for efficiently training ML algorithms. Figure 3 demonstrates the sparsity of the resulting dataset file, i.e., the data stored for features mostly contains zeroes.

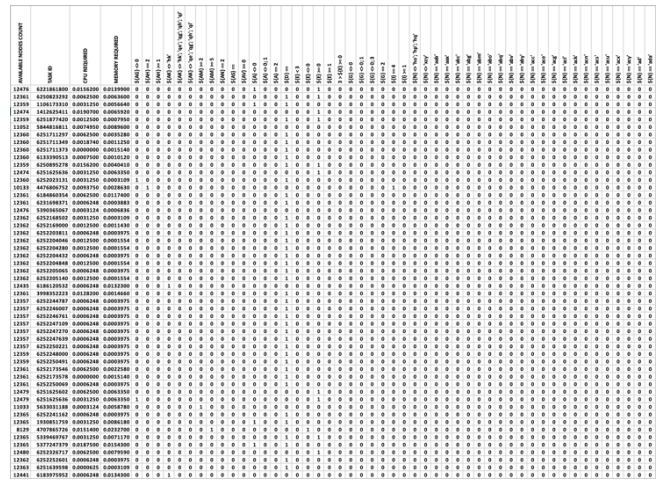

FIGURE 3. A part of the training dataset with constraint operators one-hot encoded as feature labels. Data points have more than 4.4k features.

## V. DETAILS OF MACHINE LEARNING

With the testbed data in place, the next research goal was to find a working prediction model for workload allocation difficulty. Initially, the research considered regression models, where ML algorithms predicted the number of nodes suitable for given tasks, which would then be converted into a classifying group A-Z. However, this approach proved difficult to evaluate, and only a few models returned good results. Therefore, the research switched to ML multiclass classifier models, which could handle a wide array of features and be sufficiently tuned to achieve good accuracy results, prior to an evaluation of their performance. The following sections describe the steps taken.

### A. QUANTIFYING THE QUALITY OF PREDICTIONS

During this research, the models were quantified in a range of ways using several algorithms. The models were evaluated based on general prediction accuracy and the precision of positive selection for groups A and C, i.e., the models should make the fewest number of mistakes for these two labels. The reason for this focus is that misclassifications among later groups (above C) should not significantly impact task scheduling. For example, a misclassification between group P (7001 to 7500 available nodes for a given task) and group Q (7501 to 8000 available nodes for a given task) is not critical. However, a misclassification between group A (exactly one node suitable to run a given task) and group C (501 to 1000 nodes suitable) is highly significant. For this purpose, the results also include an F1 score (a measure of the harmonic mean of precision and recall) for groups A and C.

TABLE 3
EVALUATION OF ML MODELS

| Classifier algorithm / hyperparameters | Performance | Timings | Notes |
|---|---|---|---|
| **K-Nearest Neighbours** [70] (sklearn.neighbors.KNeighborsClassifier)<br><br>Non-default hyperparameters:<br>n_neighbors=3<br>weights="distance" | Accuracy: 89%<br><br>F1 scores:<br>A: 0.9086-0.9804<br>C: 0.3333-0.8571 | Train time:<br>~40ms (elapsed)<br>~30ms (CPU)<br><br>Prediction time:<br>~7000ms (elapsed)<br>~5200ms (CPU) | A higher number of neighbours (more than three) resulted in significant misclassifications for Group C. Enabling "distance" mode, in which closer neighbours of a query point have a greater influence than neighbours further away, resulted in slightly better F1 scores for both groups A and C. |
| **Decision Tree** [69] (sklearn.tree.DecisionTreeClassifier)<br><br>Non-default hyperparameters:<br>max_depth=15<br>class_weight="balanced"<br>max_features=None | Accuracy: 95%<br><br>F1 scores:<br>A: 0.9541-0.9704<br>C: 0-0.6667 | Train time:<br>~5000ms (elapsed)<br>~5000ms (CPU)<br><br>Prediction time:<br>~40ms (elapsed)<br>~40ms (CPU) | The maximum tree depth values above fifteen did not yield better results, and substantially slowed down the training process (with default unlimited maximum depth running for over one minute). The "balanced" mode, which uses the values of dependent variables to automatically adjust weights inversely proportional to class frequencies in the input data, substantially increased recall for group C, thus improving the F1 score for this group. Additionally, the best results were found by not limiting the number of features considered when looking for the best split. |
| **Random Forest** [69] (sklearn.ensemble.RandomForestClassifier)<br><br>Non-default hyperparameters:<br>n_estimators=20<br>max_depth=10<br>class_weight="balanced"<br>max_features=None | Accuracy: 96%<br><br>F1 scores:<br>A: 0.8965-0.9587<br>C: 0.4000-1 | Train time:<br>~30000ms (elapsed)<br>~29500ms (CPU)<br><br>Prediction time:<br>~500ms (elapsed)<br>~500ms (CPU) | In comparison to a single Decision Tree, using multiple (twenty) estimators helped to lower the maximum tree depth and yielded slightly better results, especially in terms of precision for group C. As with the single Decision Tree classifier, considering all features when looking for the best split resulted in better outcomes. While the training time was substantially increased, the prediction process still lasted under half a second. |
| **Gaussian Naive Bayes** [66] (sklearn.naive_bayes.GaussianNB) | Accuracy: 43%<br><br>F1 scores:<br>A: 0.9298-9385<br>C: 0.7500-1 | Train time:<br>~800ms (elapsed)<br>~800ms (CPU)<br><br>Prediction time:<br>~2000ms (elapsed)<br>~2000ms (CPU) | The SciKit-learn implementation of this algorithm did not offer viable tuning options. |
| **Complement Naive Bayes** [65] (sklearn.naive_bayes.ComplementNB)<br><br>Non-default hyperparameters:<br>alpha=0.3 | Accuracy: 92%<br>F1 scores:<br>A: 0.9199-0.9304<br>C: 0.8333-1 | Train time:<br>~5000ms (elapsed)<br>~5000ms (CPU)<br><br>Prediction time:<br>~100ms (elapsed)<br>~100ms (CPU) | While the initial results were poor (F1 score for group C around 0.6 and accuracy slightly above 70%), this model substantially benefited from tuning. After lowering the additive (Laplace/Lidstone) smoothing parameter from 1.0 to 0.3, the accuracy became acceptable, while timings remained very low. |
| **Nearest Centroid** [63] (sklearn.neighbors.NearestCentroid) | Accuracy: 87%<br><br>F1 scores:<br>A: 0.9580-0.9630<br>C: 0.4000-0.8571 | Train time:<br>~300ms (elapsed)<br>~300ms (CPU)<br><br>Prediction time:<br>~200ms (elapsed)<br>~200ms (CPU) | A simple yet surprisingly effective model for classification. No viable tuning options were found. |
| **Ridge Regression** [62] (sklearn.linear_model.RidgeClassifier)<br><br>Non-default hyperparameters:<br>alpha=0.3<br>fit_intercept=False | Accuracy:98%<br><br>F1 scores:<br>A: 0.9933-0.9953<br>C: 0.8000-0.9091 | Train time:<br>~3500ms (elapsed)<br>~4700ms (CPU)<br><br>Prediction time:<br>~200ms (elapsed)<br>~200ms (CPU) | This classifier, based on Ridge regression, has one of the best F1 scores for group A, which is critical for efficient task scheduling, and misclassified only 30 out of 3.2k group A data points from the test set. The best results were achieved by lowering the regularisation strength and skipping centring of the input data, as Ridge regression algorithms perform well on sparse data. |
| **Linear Support Vector Machine with** [64] **Stochastic Gradient Descent (SGD) training** (sklearn.linear_model.SGDClassifier)<br><br>Non-default hyperparameters:<br>fit_intercept=False<br>max_iter=100 | Accuracy: 98%<br><br>F1 scores:<br>A: 0.9862-0.9902<br>C: 0.7500-1 | Train time:<br>~40000ms (elapsed)<br>~40000ms (CPU)<br><br>Prediction time:<br>~100ms (elapsed)<br>~100ms (CPU) | Previous attempts with the SciKit-learn implementation of Support Vector Machine models (both linear and RBF-based) were ineffective, yielding unsatisfactory results. Consequently, the research pursued a Stochastic Gradient Descent training method, which proved to be more successful. Similar to Ridge Regression, omitting the centring of input data improved results, and reducing the iteration count shortened the training time. |
| **Label Propagation** [60] (sklearn.semi_supervised.LabelPropagation)<br><br>Non-default hyperparameters:<br>max_iter=300 | Accuracy: 96%<br><br>F1 scores:<br>A: 0.9755-0.9782<br>C: 0.5000-0.8000 | Train time:<br>~16000ms (elapsed)<br>~21000ms (CPU)<br><br>Prediction time:<br>~6000ms (elapsed)<br>~11000ms (CPU) | The Label Propagation implementation does not offer many tuning options. Changing the default kernel RBF function to KNN (K-nearest neighbours' kernel) resulted in worse performance. The only applicable tuning was a reduction in maximum iterations. Nevertheless, Label Propagation is another SciKit-learn model that effectively takes advantage of the multi-core feature of the testbed machine. |
| **Passive-Aggressive** [77] (sklearn.linear_model.PassiveAggressiveClassifier) Non-default hyperparameters: max_iter=300 tol=0.1 | Accuracy: 91% F1 scores: A: 0.8965-0.9882 C: 0.3333-0.8000 | Train time: ~24000ms (elapsed) ~24000ms (CPU) Prediction time: ~50ms (elapsed) ~40ms (CPU) | Online Passive-Aggressive Algorithms are lesser-known classifiers for sparse data. Classification occurs in rounds, where the algorithm predicts a label (+1 or −1) for each instance. After the prediction, the true label is revealed, and the algorithm incurs a loss based on the prediction error. The algorithm then uses this new instance-label pair to refine its prediction rule. The two possible tunings found were to reduce the number of epochs and increase the stopping criterion tolerance. |
| **Adaptive Boosting** [61] (sklearn.ensemble.AdaBoostClassifier) Non-default hyperparameters: n_estimators=25 estimator=ExtraTreeClassifier(   splitter="random",   class_weight="balanced" ) | Accuracy: 97% F1 scores: A: 0.9768-0.9809 C: 0.6667-0.8000 | Train time: ~33000ms (elapsed) ~32200ms (CPU) Prediction time: ~800ms (elapsed) ~800ms (CPU) | Adaptive Boosting is another ensemble model that returned good results with the default hyperparameters. However, when the default estimator (Decision Tree) was replaced with Extra-tree classifier (with provided non-default hyperparameters), the results substantially improved to top levels. Extra-tree classifiers differ from classic Decision Trees in their building process. Instead of seeking the optimal split for each node, Extra-trees generate random splits for randomly selected features and then choose the best among these random splits. This approach results in a highly randomised tree classifier, which is best suited for use within ensemble methods like Adaptive Boosting. |
| **Bagging** [49] (sklearn.ensemble.BaggingClassifier) Non-default hyperparameters: bootstrap=False n_estimators=20 n_jobs=-1 estimator=ExtraTreeClassifier(   splitter="random",   class_weight="balanced" ) | Accuracy: 97% F1 scores: A: 0.9736-0.9794 C: 0.3333-0.8571 | Train time: ~15000ms (elapsed) ~40000ms (CPU[1]) Prediction time: ~2500ms (elapsed) ~8000ms (CPU[1]) | This ensemble method has several names: (i) 'pasting', when random subsets of the dataset are selected as random subsets of the samples [48]; (ii) 'bagging', when the samples are drawn with a replacement [49]; (iii) 'random subspaces', when random subsets of the dataset are chosen as random subsets of the features [50]; and (iv) 'random patches', when base estimators are constructed on subsets of both samples and features [51]. As in previous tests with Adaptive Boosting, Extra Tree estimators yielded the best results. However, it was found that the number of estimators could be reduced while still maintaining satisfactory outcomes. Disabling sample replacement during drawing slightly improved the accuracy. The option to enable parallelism is rare among tested models and using all available cores on the testbed machine effectively halved both the training and prediction time. |
| **Artificial Neural Network** [67] (sklearn.neural_network.MLPClassifier) Non-default hyperparameters: solver='adam' hidden_layer_sizes=(30,30,) max_iter=200 | Accuracy: 97% F1 scores: A: 0.9749-0.9826 C: 0.8000-1 | Train time: ~90000ms (elapsed) ~160000ms (CPU) Prediction time: ~100ms (elapsed) ~100ms (CPU) | The artificial neural network model yielded good results with the default hyperparameters, and the results were further improved after the tuning process. Initially, creating three layers and simultaneously decreasing the size of the hidden layers from the default 100 to 30 improved the measured F1 scores, albeit at the cost of increased training time. However, subsequently lowering the number of epochs for the default ADAM [68] solver substantially reduced the training time to a minute and a half without reducing the quality of results. |

1. Python's 'process_time' function was not capturing the timings correctly. The provided times were estimated with the help of the psutil tool.

## B. MODEL SELECTION

This research aimed to test a broad spectrum of ML classifier models with a particular focus on Artificial Neural Network. However, not every model was able to work with the training dataset, and the selection of algorithms was constrained by:
- the requirement that classifiers support multiple classes;
- the ability to process a dataset with 4.4k features;
- the ability to handle highly sparse data;
- the need to predict results with high accuracy, 95% or higher;
- the need to consistently minimise misclassifications for Groups A and C;
- the need to be relatively lightweight and have reasonable training and prediction times, less than five minutes;

Table 3 lists the ML models that were found to be suitable, considering the above constraints. It also summarises the research with notes on the performance of the suitable algorithms. The timings for the tested algorithms are provided for both elapsed time and CPU time. In cases where the model takes advantage of the multi-core testbed machine, the CPU time will be higher. For the purpose of model evaluation, the dataset was split into a training set of 20.7k data points and a test set of 6.9k data points. Each model was run ten times, and the presented accuracy is an average, while the F1 scores report the minimum and maximum values achieved. Where hyperparameters differ from the default values, this is explicitly mentioned in the description.

A set of other classifiers were tried, such as Gaussian Process (based on Laplace approximation), Bernoulli Restricted Boltzmann Machine, Linear Discriminant Analysis, Quadratic Discriminant Analysis, and Support Vector Machines with different kernel functions, etc. However, their results were not satisfactory or the models' timings were too long, i.e., the time required to train a model or predict results was not practical. A brief analysis of the code revealed that these models were generally not optimised for datasets with a large number of features (such as the 4.4k features in the training dataset).

## C. ARTIFICIAL NEURAL NETWORK CLASSIFIER

In this investigation, special focus was given to the Artificial Neural Network (ANN) classifier model. The ANN model can be traced back to the Perceptron algorithm, also known as the McCulloch–Pitts neuron, which dates back to 1958 and is credited to Frank Rosenblatt, who subsequently custom-built the Mark 1 Perceptron machine for pattern recognition tasks [47]. However, the introduction of hidden layers allows the network to exhibit non-linear behaviour, which is a desired capability for solving the presented problem.

A key drawback of the ANN classifier is its longer training time compared to other algorithms; nevertheless, neural networks are among the few algorithms that efficiently utilise the multiple CPU cores of the testbed machine. Unfortunately, this implementation of neural network models is not suitable for large-scale applications. Specifically, SciKit-learn lacks GPU support, and another framework, such as PyTorch or TensorFlow, is needed to fully utilise the available cores and reduce processing times. Based on available benchmarks [71][72], the processing time could be reduced by approximately three to five times.

There is a number of publications containing guidelines [73][74][75][76] on how to efficiently determine the number of hidden layers and their sizes. As a rule of thumb, to ensure the network's ability to generalise, the number of nodes should be kept as low as possible. An excess of nodes results in the neural network becoming a memory bank that can recall the training set perfectly but does not perform well on samples that were not part of the training set.

A single-layer ANN was not performant enough, resulting in an accuracy of 90% and substantial misclassifications for Group C, suggesting a non-linear relationship with class labels. Therefore, the number of layers was increased to two, each with 30 neurons. With this configuration, the ANN classifier achieved 97% accuracy and good F1 scores for both Groups A and C.

### D. PREDICTION RESULTS

Although not all the models were initially performant, after applying hyperparameters tuning (as detailed in Table 3), almost all the models performed well in terms of general accuracy.

However, as noted above, the focus of testing was to isolate models that can predict groups A and C in the most reliable manner. This is understood as having both high precision and recall, as evident in the F1-score, in particular for group A. In these terms, the following three classifier models performed the best, showing consistently high accuracy and good F1 scores for Groups A and C: Artificial Neural Network, Ridge Regression, and Linear Support Vector Machine with SGD training. Additionally, Ridge Regression had a shorter training time compared to the other two. The Adaptive Boosting model configured with the Extra Tree Classifier was shortlisted due to its high accuracy. However, the F1 scores were inconsistent, particularly for Group C. The high volatility of this estimator is inherent by design, and increasing the number of estimators did not lead to more consistent results. Therefore, the Adaptive Boosting model was removed from consideration.

To assess the performance of the three listed models, detailed confusion matrices were generated and are presented in Figures 4, 5, and 6, corresponding to the Artificial Neural Network, Ridge Regression, and Linear Support Vector Machine with Stochastic Gradient Descent (SGD) training, respectively. In these matrices, the y-axis denotes the actual values for Groups A through Z (for instance, 'true:A' for Group A), while the x-axis indicates the values predicted by the model (such as 'pred:Z' for Group Z). Consequently, the figures display counts along the diagonal that signify the number of correct predictions, whereas the off-diagonal entries reflect the number of incorrect predictions. It is important to note that any groups not predicted by the machine learning model are omitted from the matrices. Furthermore, predictions for Group A, both the correct classifications and the misclassifications, are emphasised in red for clarity.

As can be observed from the confusion matrices, while the overall accuracy is adequate, all three models occasionally misclassify group A (a task which can be allocated to exactly one node) as group Z (a task which can be allocated to virtually any node). Between 1% and 4% of tasks are misclassified in this way – this scenario is especially unfavourable as the cluster scheduler would not consider a given task to be hard to allocate.

### A. FINAL ENSEMBLE VOTING MODEL

To further remedy the results, an attempt was made with use of ensemble voting model as presented in the Listing 3. The three best models from previous steps were set up as estimators: Artificial Neural Network, Ridge Regression, and Linear Support Vector Machine with SGD training.

FIGURE 4. The confusion metrics for Artificial Neural Network

FIGURE 5. The confusion metrics for Ridge Regression

FIGURE 6. The confusion metrics for Linear Support Vector Machine with SGD training

Not all of the deployed models were equipped to retrieve class probabilities, as their implementations did not include the 'predict_proba' method. Consequently, this limitation meant that only hard voting—essentially, majority rule voting—was feasible for ensemble decision-making. Despite the use of this ensemble voting approach, which aggregated predictions from multiple models, there was no discernible improvement in terms of accuracy, which remained at 98%, or in the F1-scores, which achieved values between 0.9880 and 0.9902 for Group A, and between 0.8571 and 1.0000 for Group C. Nonetheless, the ensemble voting model did yield results that were more consistent compared to the individual models. However, this consistency came at the expense of an increased training time, which extended by approximately one minute. Tasks from Group A were misclassified at a rate of about 1.5% to 1.8%. Despite subsequent attempts to refine the model by employing alternative approaches, no additional improvements in performance were observed.

LISTING 3
FINAL ENSEMBLE VOTING MODEL

```
import numpy as np
import pandas as pd

from sklearn.ensemble import VotingClassifier
from sklearn.linear_model import RidgeClassifier, SGDClassifier
from sklearn.neural_network import MLPClassifier
from sklearn.model_selection import train_test_split
from sklearn.metrics import classification_report, accuracy_score

# load data points
dataset = pd.read_csv(f'datapoint-tasks-full.csv')
X = dataset.iloc[:, 4:].values  # columns 4+ contain features
y = dataset.iloc[:, 0].values   # nodes counts' are in the first column

X_test, X_train, y_test, y_train = train_test_split(X, y, test_size=0.75)

# group classification function
def get_classification (nodes_count):
    if nodes_count == 1: return 'A'
    elif nodes_count > 12000: return 'Z'
    return chr((int(nodes_count) - 1) // 500 + 66)

# convert counts of nodes to classification groups
vfunc_get_classification = np.vectorize(get_classification)
y_train = vfunc_get_classification(y_train)
y_test = vfunc_get_classification(y_test)

model = VotingClassifier(
    estimators=[
        ('Neural-Network',
         MLPClassifier(hidden_layer_sizes=(30, 30),
                       max_iter=200)),
        ('Ridge-Regression',
         RidgeClassifier(alpha=0.3,
                         fit_intercept=False)),
        ('Linear SVM with SGD',
         SGDClassifier(fit_intercept=False,
                       max_iter=100))
    ],
    voting='hard',  # majority voting mode
    n_jobs=-1  # use all cores
)
print(f"Training {str(model)}...")
start = perf_counter()
startp = process_time()
model.fit(X_train, y_train)
print(f"Trained in {(perf_counter() - start) * 1000:.0f} ms (elapsed)")
print(f"Trained in {(process_time() - startp) * 1000:.0f} ms (CPU)")

print(f"Predicting {str(model)}...")
start = perf_counter()
startp = process_time()
y_pred = model.predict(X_test)
print(f"Predicted in {(perf_counter() - start) * 1000:.0f} ms (elapsed)")
print(f"Predicted in {(process_time() - startp) * 1000:.0f} ms (CPU)")

print(f"Accuracy: {accuracy_score(y_test, y_pred)}")
print(classification_report(y_test, y_pred,
                            digits=4,
                            zero_division=0,
                            labels=np.unique(y_test)))
unique_label = np.unique([y_true, y_pred])
confusion_matrix_pd = pd.DataFrame(
    confusion_matrix(y_true, y_pred, labels=unique_label),
    index=[f'true:{x}' for x in unique_label],
    columns=[f'pred:{z}' for x in unique_label]
)
print(confusion_matrix_pd.to_string())
```

## VI. CONCLUSIONS AND FUTURE RESEARCH

This research introduces a novel evaluation method for submitted tasks to cluster systems. This paper represents an attempt to improve task allocations based on analysing and modelling execution constraints, which is absent in previous research or, at best, constraints logic was simplified to checking if a given node could support a given task or not.

The research has a narrow focus, with the primary contribution being the proof of concept that ML algorithms can effectively enhance cluster scheduling strategies, particularly for tasks with defined run-environment limitations. The main strength of the research lies in the utilisation of real-world workload traces from GCD archives in both the detailed simulation and analysis phases. The novelty of the presented method lies in transforming the constraint operators into an ML-suitable dataset, and then further analysing them with regard to execution-runtime constraints.

To answer the research question, stated as: **Can the number of suitable nodes for a given task be predicted based on its constraint operators?**, a set of ML algorithms was deployed and tested, and the three best were selected based on their results, accuracy, and the ratio of misclassifications for Groups A (a task which can be allocated to exactly one node) and C (containing tasks that can be scheduled on fewer than a thousand nodes). In the last step, the Artificial Neural Network, Ridge Regression, and Linear Support Vector Machine with SGD training were combined and used as estimators for the final ensemble voting classifier. Despite attempts with many different models, the tasks from Group A were occasionally misclassified as Group Z (a task which can be allocated to virtually any node). This scenario is unfavourable, as the cluster scheduler would not consider a given task to be hard to allocate. In the best solution found, this effect occurred in around 1.5-1.8% of prediction and, while this ratio seems minuscule, real-world deployment is necessary to confirm the usefulness of the proposed solution.

Moreover, while ML models contain elements of randomness and are unlikely to be fully accurate; misclassifications could potentially be filtered by a second layer of heuristic algorithms. The hybrid ML approach has been an active area of research for a considerable time and has already been successfully deployed in a number of other works, such as [79], which explores the integration of fuzzy techniques with machine learning to handle uncertainty and improve model performance; [80], where pre-processing techniques like decomposition and auto-correlation functions are utilised to capture trends and patterns in time-series data, which are then fed into ML models; in [80], the authors proposed a fuzzy transformation to convert binary inputs and outputs into a three-class system – this transformation enabled the ML models to predict a broader diagnostic spectrum; and in [82] and [83], hybrid ML models outperformed other network security systems. Therefore, it's plausible that a hybrid ML model would further reduce misclassifications.

The primary limitation of the proposed solution is its scalability – specifically, the model's inability to dynamically incorporate new constraint operators. Each time a new constraint operator is introduced (as outlined in the compaction process in Section 4), a full retraining of the ML models is required, which can be computationally intensive. An analysis of GCD workload traces indicates that tasks executed on Google data centres generally exhibit consistent requirements in terms of constraint operators. Nevertheless, techniques exist to update a running ML model with new data samples; however, not every ML algorithm is suitable for reuse in this manner. Many prior studies focus on the incremental learning of ANNs using transfer learning techniques [85][86]; however, another method to accommodate new classes is to expand the network's capacity by adding new layers [87]. In [88], the authors present a mixed approach in which an updated network is formed using previously learned convolutional layers. The aforementioned approaches could be incorporated in the next iteration of the proposed models, particularly if the cluster workload is less homogeneous.

Another concern is the potential for overfitting, particularly with the ANN classifier model. Although the implementation is not inherently complex and the algorithm can be easily modularised. The ensemble models, such as the one deployed, are inherently less prone to overfitting. It could be potentially beneficial to include more than three ML models in the final ensemble model; however, in this research, adding additional models did not yield better or more stable results. Nevertheless, it is essential to establish a mechanism for periodically monitoring the quality of results, with special attention to Group A misclassifications. A potential solution could be a continuous feedback stream from the cluster scheduler, which would be used to periodically retrain the proposed model once the ML model's results differ significantly from the true cluster capacities. Without specific cluster and workload details, it is difficult to estimate how often the model would need to be retrained. However, considering the training times (as presented in Table 3), this process would not be extensive and could be performed in parallel without interrupting cluster operations.

The analysed GCD workload traces have tasks with stable and repetitive constraints, as well as stable node configurations; however, this might not be the case in other clusters where workloads are more hectic. For example, while it is reasonable to expect production environments to be stable and to build only from standardised and well-tested applications, the non-production environments – especially testing and development clusters – are chaotic by design. In such environments, the scheduling strategies might benefit from leaving more headroom on nodes for unexpected resource utilisation spikes, rather than packing tasks tightly. However, before implementing the proposed solution in practice, it is difficult to anticipate potential issues and solutions.

Another limitation is the complexity of features, particularly the number of categories derived from workload traces, which are subsequently converted into one-hot vectors. The simulated workload traces encompassed sixty-seven distinct node attributes with multiple categories, resulting in over 4k distinctive constraint operators. If other workload traces exhibit a higher number of categories or have a greater variety of constraint operators, the approach presented may require significant modifications or be not be suitable.

The paper presents the timings of the models used for training and predicting data. The research aimed to record accurate CPU time; however, profiling SciKit-learn is surprisingly challenging due to its combination of Python code, gcc- and Fortran-compiled binaries, and other native libraries. The py-spy tool (py-spy package version 0.3.14) was used to record the performance profiles, visualised as 'flame graphs' in Figure 7, which were then analysed using www.speedscope.app. However, the results were not accurate, and this research instead includes CPU time measured by Python's native 'process_time' function, which is process-wide and includes both system and user CPU time of the current process.

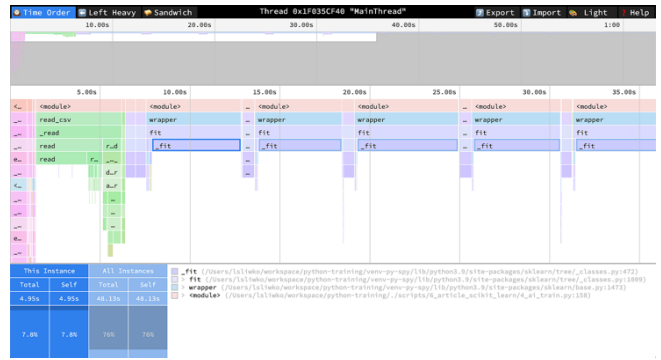

FIGURE 7. The recorded performance profile presented as a flame graph (www.speedscope.app)

Furthermore, the research presented focuses on GCD workload traces, with the method developed to handle this particular environment using custom constraint operators. However, other cluster orchestration software provides similar functionality; for example, Kubernetes offers a 'node-selector' in the pod deployment description, restricting a given pod to be scheduled only on nodes that define a certain label (i.e., a key-value pair). Moreover, since Kubernetes version 1.4, developers can use 'node-affinity' descriptors directly in a pod manifest, which provides a more expressive syntax for fine-grained control of how pods are scheduled to specific nodes.

The findings of this research demonstrate that the introduction of ML models into cluster scheduling systems is a feasible goal that could improve the performance of workload allocations. However, the research has also shown that the details of ML implementation are crucial, and a real-world scenario would need to utilise more advanced and specialised frameworks such as PyTorch, TensorFlow, etc.

The investigation has highlighted a few potential directions for further research:
- Enhancing existing scheduler algorithms – the proof-of-concept implementations (as demonstrated in this paper) are feasible, and the created models can be deployed to enhance existing cluster schedulers, enabling the utilisation of alternative strategies for handling detected hard-to-allocate tasks. Therefore, the next logical step is to attempt to modify task scheduling strategies in existing open-source frameworks and test them in real-world scenarios. A promising candidate for testing these changes is Kubernetes due to its well-documented source code and modular architecture, though this requires further analysis;
- The misclassification of a single-node task as a multi-node task is certainly a significant fault, but it is not catastrophic in itself. Modern cluster schedulers are equipped with mechanisms designed to handle such situations, and they can reallocate the workload dynamically across the cluster to ensure continuity. While these schedulers can resolve the issue, the process comes with a cost, and it might involve forcibly terminating or otherwise removing other running tasks and services to free up resources, which can negatively impact overall system performance. To reduce these inefficiencies, an enhanced model could be developed to include a secondary layer of heuristics. These additional heuristics would aim to capture and address edge scenarios more effectively, potentially preventing costly misclassifications before they escalate into larger system-level problems;
- Exploring the concept of dynamically adjusting ML models that can accommodate new constraint operands without the need for fully retraining the model. This appears achievable through dynamically evolving deep neural networks [78], although further research is required. To implement such low-level changes, SciKit-learn may not be the best framework; instead, more advanced frameworks like PyTorch should be considered for this research;
- This research would benefit from experimentation on additional cluster work traces, especially from sources other than Google. However, this is not a trivial task, as other traces - such as those from the Grid Workload Archive, MetaCentrum, or Los Alamos National Laboratory's supercomputers, etc., are often not detailed enough for simulation purposes or lack node-affinity settings for tasks. Alternative sources of work traces include private organisations where Kubernetes clusters are widely used; however, these companies would need to be approached, and the necessary permissions obtained;
- Node 'soft' affinity – in many cases, a developer will use 'node-selector' to deploy pods on nodes with specific access control credentials, designated access to ports, existing licenses for third-party services, security parameters, and physical isolation. Consequently, the constraint operators establish 'hard' affinity, meaning the pod can only run if there are available nodes in the cluster that satisfy all the given criteria. However, Kubernetes has also introduced support for 'soft' node affinity [8]. 'Soft' affinity functions as a preference; for instance, with 'soft' affinity, a scheduler can be instructed to attempt to run the set of pods in a defined Availability Zone. If this proves unfeasible, it may allow some of these pods to run in the other Availability Zone. Such division between 'hard' and 'soft' node affinity creates another complexity layer which should be considered in the continuation of this research.

## ACKNOWLEDGMENT

The author of this paper would like to thank Google engineers, and in particular John Wilkes, for describing the internal workings of the Borg scheduler and enabling access to detailed Google Cluster Data workload traces.

## REFERENCES

1. Dubey, Parul, and Arvind Kumar Tiwari. "AWS Spot Instances: A Cloud Computing Cost Investigation Across AWS Regions." International Journal of Intelligent Systems and Applications in Engineering 12, no. 1 (2024): 01-10.
2. Verma, Abhishek, Luis Pedrosa, Madhukar Korupolu, David Oppenheimer, Eric Tune, and John Wilkes. "Large-scale cluster management at Google with Borg." In Proceedings of the Tenth European Conference on Computer Systems, p. 18. ACM, 2015.
3. Wilkes, John. "Cluster Management at Google with Borg." GOTO Berlin 2016. November 15, 2016.
4. Hellerstein, Joseph L., W. Cirne, and J. Wilkes. "Google Cluster Data." Google Research Blog. January 7, 2010.
5. Boutin, Eric, Jaliya Ekanayake, Wei Lin, Bing Shi, Jingren Zhou, Zhengping Qian, Ming Wu, and Lidong Zhou. "Apollo: Scalable and Coordinated Scheduling for Cloud-Scale Computing." In *OSDI*, vol. 14, pp. 285-300. 2014.
6. Zhang, Zhuo, Chao Li, Yangyu Tao, Renyu Yang, Hong Tang, and Jie Xu. "Fuxi: a fault-tolerant resource management and job scheduling system at internet scale." Proceedings of the VLDB Endowment 7, no. 13 (2014): 1393-1404.
7. Naik, Nitin. "Building a virtual system of systems using Docker Swarm in multiple clouds." In Systems Engineering (ISSE), 2016 IEEE International Symposium on, pp. 1-3. IEEE, 2016.
8. Sayfan, Gigi. Mastering Kubernetes: Level up your container orchestration skills with Kubernetes to build, run, secure, and observe large-scale distributed apps. Packt Publishing Ltd, 2020.
9. Alhilali, Ahmed Hazim, and Ahmadreza Montazerolghaem. "Artificial intelligence based load balancing in SDN: A comprehensive survey." Internet of Things (2023): 100814.
10. Kumar, Jitendra, and Ashutosh Kumar Singh. "Workload prediction in cloud using artificial neural network and adaptive differential evolution." *Future Generation Computer Systems* 81 (2018): 41-52.
11. Wilson Prakash, S., and P. Deepalakshmi. "Artificial neural network based load balancing on software defined networking." In *2019 IEEE International Conference on Intelligent Techniques in Control, Optimization and Signal Processing (INCOS)*, pp. 1-4. IEEE, 2019.
12. Melnik, Mikhail, and Denis Nasonov. "Workflow scheduling using neural networks and reinforcement learning." *Procedia computer science* 156 (2019): 29-36.
13. Kumar, Jitendra, Rimsha Goomer, and Ashutosh Kumar Singh. "Long short term memory recurrent neural network (LSTM-RNN) based workload forecasting model for cloud datacenters." *Procedia computer science* 125 (2018): 676-682.
14. Calzarossa, Maria Carla, Luisa Massari, and Daniele Tessera. "Workload characterization: A survey revisited." *ACM Computing Surveys (CSUR)* 48, no. 3 (2016): 1-43.

15. Zhou, Guangyao, Wenhong Tian, Rajkumar Buyya, Ruini Xue, and Liang Song. "Deep reinforcement learning-based methods for resource scheduling in cloud computing: A review and future directions." *Artificial Intelligence Review* 57, no. 5 (2024): 124.
16. Duc, Thang Le, Rafael García Leiva, Paolo Casari, and Per-Olov Östberg. "Machine learning methods for reliable resource provisioning in edge-cloud computing: A survey." *ACM Computing Surveys (CSUR)* 52, no. 5 (2019): 1-39.
17. Herbst, Nikolas, Ayman Amin, Artur Andrzejak, Lars Grunske, Samuel Kounev, Ole J. Mengshoel, and Priya Sundararajan. "Online workload forecasting." *Self-Aware Computing Systems* (2017): 529-553.
18. Muchori, Juliet G., and Peter M. Mwangi. "Machine Learning Load Balancing Techniques in Cloud Computing: A Review." (2022).
19. Kaur, Amanpreet, Bikrampal Kaur, Parminder Singh, Mandeep Singh Devgan, and Harpreet Kaur Toor. "Load balancing optimization based on deep learning approach in cloud environment." *International Journal of Information Technology and Computer Science* 12, no. 3 (2020): 8-18.
20. Zhu, Xiaoke, Qi Zhang, Taining Cheng, Ling Liu, Wei Zhou, and Jing He. "DLB: deep learning based load balancing." In *2021 IEEE 14th International Conference on Cloud Computing (CLOUD)*, pp. 648-653. IEEE, 2021.
21. Lilhore, Umesh Kumar, Sarita Simaiya, Kalpna Guleria, and Devendra Prasad. "An efficient load balancing method by using machine learning-based VM distribution and dynamic resource mapping." *Journal of Computational and Theoretical Nanoscience* 17, no. 6 (2020): 2545-2551.
22. Liang, Siyuan, Wenli Jiang, Fangli Zhao, and Feng Zhao. "Load balancing algorithm of controller based on sdn architecture under machine learning." *Journal of Systems Science and Information* 8, no. 6 (2020): 578-588.
23. Abdennebi, Anes, Anıl Elakaş, Fatih Taşyaran, Erdinç Öztürk, Kamer Kaya, and Sinan Yıldırım. "Machine learning-based load distribution and balancing in heterogeneous database management systems." *Concurrency and Computation: Practice and Experience* 34, no. 4 (2022): e6641.
24. Oikawa, CR Anna Victoria, Vinicius Freitas, Márcio Castro, and Laércio L. Pilla. "Adaptive load balancing based on machine learning for iterative parallel applications." In *2020 28th Euromicro International Conference on Parallel, Distributed and Network-Based Processing (PDP)*, pp. 94-101. IEEE, 2020.
25. John Wilkes, "Yet More Google Compute Cluster Trace Data", Google Research Blog, https://research.google/blog/yet-more-google-compute-cluster-trace-data/, April 28, 2020.
26. Iosup, Alexandru, Hui Li, Mathieu Jan, Shanny Anoep, Catalin Dumitrescu, Lex Wolters, and Dick HJ Epema. "The grid workloads archive." Future Generation Computer Systems 24, no. 7 (2008): 672-686.
27. Feitelson, Dror G., Dan Tsafrir, and David Krakov. "Experience with using the parallel workloads archive." Journal of Parallel and Distributed Computing 74, no. 10 (2014): 2967-2982.
28. Klusáček, Dalibor, and Boris Parák. "Analysis of Mixed Workloads from Shared Cloud Infrastructure." In Workshop on Job Scheduling Strategies for Parallel Processing, pp. 25-42. Springer, Cham, 2017.
29. Kavulya, Soila, Jiaqi Tan, Rajeev Gandhi, and Priya Narasimhan. "An analysis of traces from a production mapreduce cluster." In Proceedings of the 2010 10th IEEE/ACM International Conference on Cluster, Cloud and Grid Computing, pp. 94-103. IEEE Computer Society, 2010.
30. Schwarzkopf, Malte, Andy Konwinski, Michael Abd-El-Malek, and John Wilkes. "Omega: flexible, scalable schedulers for large compute clusters." In Proceedings of the 8th ACM European Conference on Computer Systems, pp. 351-364. ACM, 2013.
31. Chen, Yanpei, Sara Alspaugh, and Randy H. Katz. Design insights for MapReduce from diverse production workloads. No. UCB/EECS-2012-17. California Unversity Berkley, Department of Electrical Engineering and Computer Science, 2012.
32. Mishra, Asit K., Joseph L. Hellerstein, Walfredo Cirne, and Chita R. Das. "Towards characterizing cloud backend workloads: insights from Google compute clusters." ACM SIGMETRICS Performance Evaluation Review 37, no. 4 (2010): 34-41.
33. Sharma, Bikash, Victor Chudnovsky, Joseph L. Hellerstein, Rasekh Rifaat, and Chita R. Das. "Modeling and synthesizing task placement constraints in Google compute clusters." In Proceedings of the 2nd ACM Symposium on Cloud Computing, p. 3. ACM, 2011.
34. Zhang, Qi, Joseph L. Hellerstein, and Raouf Boutaba. "Characterizing task usage shapes in Google's compute clusters." In Proceedings of the 5th international workshop on large scale distributed systems and middleware, pp. 1-6. sn, 2011.
35. Reiss, Charles, John Wilkes, and Joseph L. Hellerstein. "Google cluster-usage traces: format+schema." Google, Inc. Version of 2013.05.06, for trace version 2, 2013.
36. Sliwko, Leszek, and Vladimir Getov. "AGOCS - accurate google cloud simulator framework." In 2016 Intl IEEE Conferences on Ubiquitous Intelligence & Computing, Advanced and Trusted Computing, Scalable Computing and Communications, Cloud and Big Data Computing, Internet of People, and Smart World Congress (UIC/ATC/ScalCom/CBDCom/IoP/SmartWorld), pp. 550-558. IEEE, 2016.
37. Carvalho, Marcus, Walfredo Cirne, Francisco Brasileiro, and John Wilkes. "Long-term SLOs for reclaimed cloud computing resources." In *Proceedings of the ACM Symposium on Cloud Computing*, pp. 1-13. 2014.
38. Sliwko, Leszek, and Vladimir Getov. "A meta-heuristic load balancer for cloud computing systems." In 2015 IEEE 39th Annual Computer Software and Applications Conference, vol. 3, pp. 121-126. IEEE, 2015.
39. Leszek, Sliwko. "A scalable service allocation negotiation for cloud computing." Journal of Theoretical and Applied Information Technology 96, no. 20 (2018): 6751-6782.
40. Géron, Aurélien. Hands-on machine learning with Scikit-Learn, Keras, and TensorFlow. " O'Reilly Media, Inc.", 2022.
41. McKinney, Wes. "pandas: a foundational Python library for data analysis and statistics." *Python for high performance and scientific computing* 14, no. 9 (2011): 1-9.
42. Harris, Charles R., K. Jarrod Millman, Stéfan J. Van Der Walt, Ralf Gommers, Pauli Virtanen, David Cournapeau, Eric Wieser et al. "Array programming with NumPy." Nature 585, no. 7825 (2020): 357-362.
43. Nakata, Maho. "Basics and Practice of Linear Algebra Calculation Library BLAS and LAPACK." The Art of High Performance Computing for Computational Science, Vol. 1: Techniques of Speedup and Parallelization for General Purposes (2019): 83-112.
44. Accelerate | Apple Developer Documentation, Apple Inc., https://developer.apple.com/documentation/accelerate. Retrieved May 23, 2024.
45. Feitelson, Dror G., and Larry Rudolph. "Gang scheduling performance benefits for fine-grain synchronization." *Journal of Parallel and distributed Computing* 16, no. 4 (1992): 306-318.
46. Jajoo, Akshay, Y. Charlie Hu, Xiaojun Lin, and Nan Deng. "A case for task sampling based learning for cluster job scheduling." In *19th USENIX Symposium on Networked Systems Design and Implementation (NSDI 22)*, pp. 19-33. 2022.
47. Rosenblatt, Frank. "The perceptron: a probabilistic model for information storage and organization in the brain." *Psychological review* 65, no. 6 (1958): 386.
48. L. Breiman, "Pasting small votes for classification in large databases and on-line", Machine Learning, 36(1), 85-103, 1999.
49. L. Breiman, "Bagging predictors", Machine Learning, 24(2), 123-140, 1996.
50. T. Ho, "The random subspace method for constructing decision forests", Pattern Analysis and Machine Intelligence, 20(8), 832-844, 1998.
51. G. Louppe and P. Geurts, "Ensembles on Random Patches", Machine Learning and Knowledge Discovery in Databases, 346-361, 2012.
52. Simaiya, Sarita, Umesh Kumar Lilhore, Yogesh Kumar Sharma, KBV Brahma Rao, V. V. R. Maheswara Rao, Anupam Baliyan, Anchit Bijalwan, and Roobaea Alroobaea. "A hybrid cloud load balancing and host utilization prediction method using deep learning and optimization techniques." *Scientific Reports* 14, no. 1 (2024): 1337.
53. Kadakia, Dharmesh. Apache Mesos Essentials. Packt Publishing, 2015.
54. Soppelsa, Fabrizio, and Chanwit Kaewkasi. Native docker clustering with swarm. Packt Publishing Ltd, 2016.
55. Seger, Cedric. "An investigation of categorical variable encoding techniques in machine learning: binary versus one-hot and feature hashing." (2018).
56. Xu, Zheng, Yulu Gong, Yanlin Zhou, Qiaozhi Bao, and Wenpin Qian. "Enhancing Kubernetes Automated Scheduling with Deep Learning and Reinforcement Techniques for Large-Scale Cloud Computing Optimization." arXiv preprint arXiv:2403.07905 (2024).
57. Burns, Brendan, Brian Grant, David Oppenheimer, Eric Brewer, and John Wilkes. "Borg, Omega, and Kubernetes." Communications of the ACM 59, no. 5 (2016): 50-57.
58. Amir, Yair, Baruch Awerbuch, Amnon Barak, R. Sean Borgstrom, and Arie Keren. "An opportunity cost approach for job assignment in a scalable computing cluster." IEEE Transactions on parallel and distributed Systems 11, no. 7 (2000): 760-768.
59. Wang, Xin, Kai Zhao, and Bin Qin. "Optimization of Task-Scheduling Strategy in Edge Kubernetes Clusters Based on Deep Reinforcement Learning." Mathematics 11, no. 20 (2023): 4269.
60. Zhu, Xiaojin, and Zoubin Ghahramani. "Learning from labeled and unlabeled data with label propagation." ProQuest number: information to all users (2002).
61. Hastie, Trevor, Saharon Rosset, Ji Zhu, and Hui Zou. "Multi-class adaboost." *Statistics and its Interface* 2, no. 3 (2009): 349-360.
62. Hoerl, Arthur E., and Robert W. Kennard. "Ridge regression: Biased estimation for nonorthogonal problems." *Technometrics* 12, no. 1 (1970): 55-67.

63. Tibshirani, Robert, Trevor Hastie, Balasubramanian Narasimhan, and Gilbert Chu. "Diagnosis of multiple cancer types by shrunken centroids of gene expression." *Proceedings of the National Academy of Sciences* 99, no. 10 (2002): 6567-6572.
64. Zhang, Tong. "Solving large scale linear prediction problems using stochastic gradient descent algorithms." In *Proceedings of the twenty-first international conference on Machine learning*, p. 116. 2004.
65. Rennie, Jason D., Lawrence Shih, Jaime Teevan, and David R. Karger. "Tackling the poor assumptions of naive bayes text classifiers." In *Proceedings of the 20th international conference on machine learning (ICML-03)*, pp. 616-623. 2003.
66. Hastie, Trevor, Robert Tibshirani, Jerome H. Friedman, and Jerome H. Friedman. The elements of statistical learning: data mining, inference, and prediction. Vol. 2. New York: springer, 2009.
67. Glorot, Xavier, and Yoshua Bengio. "Understanding the difficulty of training deep feedforward neural networks." In Proceedings of the thirteenth international conference on artificial intelligence and statistics, pp. 249-256. JMLR Workshop and Conference Proceedings, 2010.
68. Kingma, Diederik P., and Jimmy Ba. "Adam: A method for stochastic optimization." arXiv preprint arXiv:1412.6980 (2014).
69. Breiman, Leo. "Random forests." Machine learning 45 (2001): 5-32.
70. Cover, Thomas, and Peter Hart. "Nearest neighbor pattern classification." IEEE transactions on information theory 13, no. 1 (1967): 21-27.
71. Vinil Vadakkepurakkal, "Exploring CPU vs GPU Speed in AI Training: A Demonstration with TensorFlow", Azure High Performance Computing (HPC) Blog, https://techcommunity.microsoft.com/t5/azure-high-performance-computing/exploring-cpu-vs-gpu-speed-in-ai-training-a-demonstration-with/ba-p/4014242, December 20, 2023.
72. Özgür, Atilla, and Hamit Erdem. "The impact of using large training data set KDD99 on classification accuracy." PeerJ PrePrints 5 (2017): e2838v1.
73. Yang, Tien-Ju, and Vivienne Sze. "Design considerations for efficient deep neural networks on processing-in-memory accelerators." In 2019 IEEE International Electron Devices Meeting (IEDM), pp. 22-1. IEEE, 2019.
74. Smithson, Sean C., Guang Yang, Warren J. Gross, and Brett H. Meyer. "Neural networks designing neural networks: multi-objective hyper-parameter optimization." In 2016 IEEE/ACM International Conference on Computer-Aided Design (ICCAD), pp. 1-8. IEEE, 2016.
75. Hagan, Martin T., Howard B. Demuth, and Mark Beale. Neural network design. PWS Publishing Co., 1997.
76. Sheela, K. Gnana, and Subramaniam N. Deepa. "Review on methods to fix number of hidden neurons in neural networks." Mathematical problems in engineering 2013, no. 1 (2013): 425740.
77. Crammer, Koby, Ofer Dekel, Joseph Keshet, Shai Shalev-Shwartz, Yoram Singer, and Manfred K. Warmuth. "Online passive-aggressive algorithms." Journal of Machine Learning Research 7, no. 3 (2006).
78. Zhong, Yuan, Jing Zhou, Ping Li, and Jie Gong. "Dynamically evolving deep neural networks with continuous online learning." Information Sciences 646 (2023): 119411.
79. Lu, Jie, Guangzhi Ma, and Guangquan Zhang. "Fuzzy Machine Learning: A Comprehensive Framework and Systematic Review." IEEE Transactions on Fuzzy Systems (2024).
80. Liu, Shun, Kexin Wu, Chufeng Jiang, Bin Huang, and Danqing Ma. "Financial time-series forecasting: Towards synergizing performance and interpretability within a hybrid machine learning approach." arXiv preprint arXiv:2401.00534 (2023).
81. Khushal, Rabia, and Ubaida Fatima. "Fuzzy machine learning logic utilization on hormonal imbalance dataset." Computers in Biology and Medicine 174 (2024): 108429.
82. Talukder, Md Alamin, Khondokar Fida Hasan, Md Manowarul Islam, Md Ashraf Uddin, Arnisha Akhter, Mohammand Abu Yousuf, Fares Alharbi, and Mohammad Ali Moni. "A dependable hybrid machine learning model for network intrusion detection." Journal of Information Security and Applications 72 (2023): 103405.
83. Bansal, Vipul, Ankit Bhardwaj, Jitendra Singh, Devvret Verma, Mohit Tiwari, and Someshwar Siddi. "Using artificial intelligence to integrate machine learning, fuzzy logic, and the IOT as A cybersecurity system." In 2023 3rd International Conference on Advance Computing and Innovative Technologies in Engineering (ICACITE), pp. 762-769. IEEE, 2023.
84. Tirmazi, Muhammad, Adam Barker, Nan Deng, Md E. Haque, Zhijing Gene Qin, Steven Hand, Mor Harchol-Balter, and John Wilkes. "Borg: the next generation." In Proceedings of the fifteenth European conference on computer systems, pp. 1-14. 2020.
85. Fei-Fei, Li, Robert Fergus, and Pietro Perona. "One-shot learning of object categories." IEEE transactions on pattern analysis and machine intelligence 28, no. 4 (2006): 594-611.
86. Lampert, Christoph H., Hannes Nickisch, and Stefan Harmeling. "Learning to detect unseen object classes by between-class attribute transfer." In 2009 IEEE conference on computer vision and pattern recognition, pp. 951-958. IEEE, 2009.
87. Rusu, Andrei A., Neil C. Rabinowitz, Guillaume Desjardins, Hubert Soyer, James Kirkpatrick, Koray Kavukcuoglu, Razvan Pascanu, and Raia Hadsell. "Progressive neural networks." arXiv preprint arXiv:1606.04671 (2016).
88. Sarwar, Syed Shakib, Aayush Ankit, and Kaushik Roy. "Incremental learning in deep convolutional neural networks using partial network sharing." IEEE Access 8 (2019): 4615-4628.

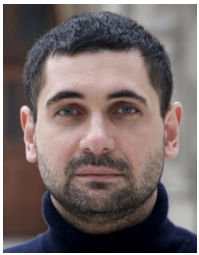

**LESZEK SLIWKO** was awarded a PhD in Parallel and Distributed Computing from the University of Westminster (London, UK) in 2019 for his work on developing a novel AI-driven load balancer for cloud systems. He has presented at international events such as IEEE conferences, Machine Intelligence Research Lab lectures, and popular seminars like Scala in the City, JVM Roundabout, and Scala Central. His talks are available at https://www.youtube.com/@lsliwko.

He is keen on experimenting with the latest technologies, particularly those related to high-performance computing and machine learning. Leszek has been an IEEE Professional member since 2019. A brief description of his selected projects can be found at https://meta-analyzer.com/projects/.